# Modeling Documents with a Deep Boltzmann Machine

**Nitish Srivastava**  **Ruslan Salakhutdinov**  **Geoffrey Hinton**
{nitish, rsalakhu, hinton}@cs.toronto.edu
Department of Computer Science, University of Toronto
Toronto, Ontario, M5S 3G4 Canada.

## Abstract

We introduce a type of Deep Boltzmann Machine (DBM) that is suitable for extracting distributed semantic representations from a large unstructured collection of documents. We overcome the apparent difficulty of training a DBM with judicious parameter tying. This enables an efficient pretraining algorithm and a state initialization scheme for fast inference. The model can be trained just as efficiently as a standard Restricted Boltzmann Machine. Our experiments show that the model assigns better log probability to unseen data than the Replicated Softmax model. Features extracted from our model outperform LDA, Replicated Softmax, and DocNADE models on document retrieval and document classification tasks.

## 1 Introduction

Text documents are a ubiquitous source of information. Representing the information content of a document in a form that is suitable for solving real-world problems is an important task. The aim of topic modeling is to create such representations by discovering latent topic structure in collections of documents. These representations are useful for document classification and retrieval tasks, making topic modeling an important machine learning problem.

The most common approach to topic modeling is to build a generative probabilistic model of the bag of words in a document. Directed graphical models, such as Latent Dirichlet Allocation (LDA), CTM, H-LDA, have been extensively used for this [3, 2, 8]. Non-parametric extensions of these models have also been quite successful [13, 1, 5]. Even though exact inference in these models is hard, efficient inference schemes, including stochastic variational inference, online inference, and collapsed Gibbs have been developed that make it feasible to train and use these methods [14, 16, 4]. Another approach is to use undirected graphical models such as the Replicated Softmax model [12]. In this model, inferring latent topic representations is exact and efficient. However, training is still hard and often requires careful hyperparameter selection. These models typically perform better than LDA in terms of both the log probability they assign to unseen data and their document retrieval and document classification accuracy. Recently, neural network based approaches, such as Neural Autoregressive Density Estimators (DocNADE) [7], have been to shown to outperform the Replicated Softmax model.

The Replicated Softmax model is a family of Restricted Boltzmann Machines (RBMs) with shared parameters. An important feature of RBMs is that they solve the "explaining-way" problem of directed graphical models by having a complementary prior over hidden units. However, this implicit prior may not be the best prior to use and having some degree of flexibility in defining the prior may be advantageous. One way of adding this additional degree of flexibility, while still avoiding the explaining-away problem, is to learn a two hidden layer Deep Boltzmann Machine (DBM). This model adds another layer of hidden units on top of the first hidden layer with bi-partite, undirected connections. The new connections come with a new set of weights. However, this additional implicit prior comes at the cost of more expensive training and inference. Therefore, we have the following two extremes: On one hand, RBMs can be efficiently trained (e.g. using Contrastive Divergence), inferring the state of the hidden units is exact, but the model defines a rigid, implicit prior. On the other hand, a two hidden layer DBM defines a more flexible prior over the hidden representations, but training and performing inference in a DBM model is considerably harder.

In this paper, we try to find middle ground between these extremes and build a model that combines the best of both. We introduce a two hidden layer DBM model, which we call the Over-Replicated Softmax model. This model is easy to train, has fast approximate inference and still retains some degree of flexibility towards manipulating the prior. Our experiments show that this flexibility is enough to improve significantly on the performance of the standard Replicated Softmax model, both as generative models and as feature extractors even though the new model only has one more parameter than the RBM model. The model also outperforms LDA and DocNADE in terms of classification and retrieval tasks.

Before we describe our model, we briefly review the Replicated Softmax model [12] which is a stepping stone towards the proposed Over-Replicated Softmax model.

## 2 Replicated Softmax Model

This model comprises of a family of Restricted Boltzmann Machines. Each RBM has "softmax" visible variables that can have one of a number of different states. Specifically, let $K$ be the dictionary size, $N$ be the number of words appearing in a document, and $\mathbf{h} \in \{0,1\}^F$ be binary stochastic hidden topic features. Let $\mathbf{V}$ be a $N \times K$ observed binary matrix with $v_{ik} = 1$ if visible unit $i$ takes on the $k^{th}$ value. We define the energy of the state $\{\mathbf{V}, \mathbf{h}\}$ as :

$$E(\mathbf{V}, \mathbf{h}; \boldsymbol{\theta}) = -\sum_{i=1}^{N}\sum_{j=1}^{F}\sum_{k=1}^{K} W_{ijk} h_j v_{ik} \quad (1)$$
$$-\sum_{i=1}^{N}\sum_{k=1}^{K} v_{ik} b_{ik} - N\sum_{j=1}^{F} h_j a_j,$$

where $\boldsymbol{\theta} = \{\mathbf{W}, \mathbf{a}, \mathbf{b}\}$ are the model parameters; $W_{ijk}$ is a symmetric interaction term between visible unit $i$ that takes on value $k$, and hidden feature $j$, $b_{ik}$ is the bias of unit $i$ that takes on value $k$, and $a_j$ is the bias of hidden feature $j$. The probability that the model assigns to a visible binary matrix $\mathbf{V}$ is:

$$P(\mathbf{V}; \boldsymbol{\theta}) = \frac{1}{\mathcal{Z}(\boldsymbol{\theta}, N)} \sum_{\mathbf{h}} \exp(-E(\mathbf{V}, \mathbf{h}; \boldsymbol{\theta})) \quad (2)$$
$$\mathcal{Z}(\boldsymbol{\theta}, N) = \sum_{\mathbf{V}'}\sum_{\mathbf{h}'} \exp(-E(\mathbf{V}', \mathbf{h}'; \boldsymbol{\theta})),$$

where $\mathcal{Z}(\boldsymbol{\theta}, N)$ is known as the partition function, or normalizing constant.

The key assumption of the Replicated Softmax model is that for each document we create a separate RBM with as many softmax units as there are words in the document, as shown in Fig. 1. Assuming that the order of the words can be ignored, all of these softmax units can share the same set of weights, connecting them to binary hidden units. In this case, the energy of the state $\{\mathbf{V}, \mathbf{h}\}$ for a document that contains $N$ words is defined as:

$$E(\mathbf{V}, \mathbf{h}) = -\sum_{j=1}^{F}\sum_{k=1}^{K} W_{jk} h_j \hat{v}_k - \sum_{k=1}^{K} \hat{v}_k b_k - N\sum_{j=1}^{F} h_j a_j,$$

where $\hat{v}_k = \sum_{i=1}^{N} v_i^k$ denotes the count for the $k^{th}$ word. The bias terms of the hidden variables are scaled up by the length of the document. This scaling is important as it allows hidden units to behave sensibly when dealing with documents of different lengths. The conditional distributions are given by softmax and logistic functions:

$$P(h_j^{(1)} = 1) = \sigma\left(\sum_{k=1}^{K} W_{jk}\hat{v}_k + Na_j\right), \quad (3)$$

$$P(v_{ik} = 1) = \frac{\exp\left(\sum_{j=1}^{F} W_{jk} h_j^{(1)} + b_k\right)}{\sum_{k'=1}^{K} \exp\left(\sum_{j=1}^{F} W_{jk'} h_j^{(1)} + b_{k'}\right)}. \quad (4)$$

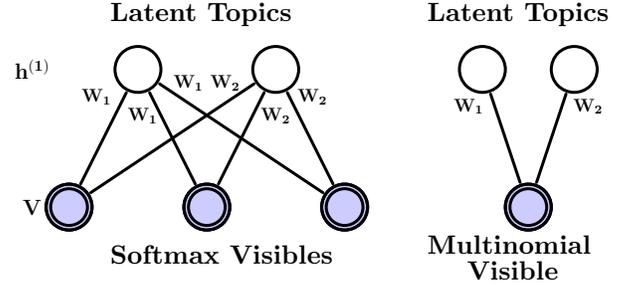

Figure 1: The Replicated Softmax model. The top layer represents a vector $\mathbf{h}$ of stochastic, binary topic features and the bottom layer represents softmax visible units $\mathbf{V}$. All visible units share the same set of weights, connecting them to binary hidden units. **Left:** The model for a document containing three words. **Right:** A different interpretation of the Replicated Softmax model, in which $N$ softmax units with identical weights are replaced by a single multinomial unit which is sampled $N$ times.

The Replicated Softmax model can also be interpreted as an RBM model that uses a single visible multinomial unit with support $\{1, ..., K\}$ which is sampled $N$ times (see Fig. 1, right panel).

For this model, exact maximum likelihood learning is intractable, because computing the derivatives of the partition function, needed for learning, takes time that is exponential in $\min\{D, F\}$, i.e the number of visible or hidden units. In practice, approximate learning is performed using Contrastive Divergence (CD) [6].

## 3 Over-Replicated Softmax Model

The Over-Replicated Softmax model is a family of two hidden layer Deep Boltzmann Machines (DBM). Let

us consider constructing a Boltzmann Machine with two hidden layers for a document containing $N$ words, as shown in Fig. 2. The visible layer $\mathbf{V}$ consists of $N$ softmax units. These units are connected to a binary hidden layer $\mathbf{h}^{(1)}$ with shared weights, exactly like in the Replicated Softmax model in Fig. 1. The second hidden layer consists of $M$ softmax units represented by $\mathbf{H}^{(2)}$. Similar to $\mathbf{V}$, $\mathbf{H}^{(2)}$ is an $M \times K$ binary matrix with $h_{mk}^{(2)} = 1$ if the $m$-th hidden softmax unit takes on the $k$-th value.

The energy of the joint configuration $\{\mathbf{V}, \mathbf{h}^{(1)}, \mathbf{H}^{(2)}\}$ is defined as:

$$E(\mathbf{V}, \mathbf{h}^{(1)}, \mathbf{H}^{(2)}; \boldsymbol{\theta}) = -\sum_{i=1}^{N}\sum_{j=1}^{F}\sum_{k=1}^{K} W_{ijk}^{(1)} h_j^{(1)} v_{ik} \quad (5)$$
$$-\sum_{i'=1}^{M}\sum_{j=1}^{F}\sum_{k=1}^{K} W_{i'jk}^{(2)} h_j^{(1)} h_{i'k}^{(2)} - \sum_{i=1}^{N}\sum_{k=1}^{K} v_{ik} b_{ik}^{(1)}$$
$$-(M+N)\sum_{j=1}^{F} h_j^{(1)} a_j - \sum_{i=1}^{M}\sum_{k=1}^{K} h_{ik}^{(2)} b_{ik}^{(2)}$$

where $\boldsymbol{\theta} = \{\mathbf{W}^{(1)}, \mathbf{W}^{(2)}, \mathbf{a}, \mathbf{b}^{(1)}, \mathbf{b}^{(2)}\}$ are the model parameters.

Similar to the Replicated Softmax model, we create a separate document-specific DBM with as many visible softmax units as there are words in the document. We also fix the number $M$ of the second-layer softmax units across all documents. Ignoring the order of the words, all of the first layer softmax units share the same set of weights. Moreover, the first and second layer weights are tied. Thus we have $W_{ijk}^{(1)} = W_{i'jk}^{(2)} = W_{jk}$ and $b_{ik}^{(1)} = b_{i'k}^{(2)} = b_k$. Compared to the standard Replicated Softmax model, this model has more replicated softmaxes (hence the name "Over-Replicated"). Unlike the visible softmaxes, these additional softmaxes are unobserved and constitute a second hidden layer[1]. The energy can be simplified to:

$$E(\mathbf{V}, \mathbf{h}^{(1)}, \mathbf{H}^{(2)}; \boldsymbol{\theta}) = -\sum_{j=1}^{F}\sum_{k=1}^{K} W_{jk} h_j^{(1)} \left(\hat{v}_k + \hat{h}_k^{(2)}\right) \quad (6)$$
$$-\sum_{k=1}^{K} \left(\hat{v}_k + \hat{h}_k^{(2)}\right) b_k - (M+N)\sum_{j=1}^{F} h_j^{(1)} a_j$$

where $\hat{v}_k = \sum_{i=1}^{N} v_{ik}$ denotes the count for the $k^{th}$ word in the input and $\hat{h}_k^{(2)} = \sum_{i=1}^{M} h_{ik}^{(2)}$ denotes the count for the $k^{th}$ "latent" word in the second hidden layer. The joint probability distribution is defined as:

$$P(\mathbf{V}, \mathbf{h}^{(1)}, \mathbf{H}^{(2)}; \boldsymbol{\theta}) = \frac{\exp\left(-E(\mathbf{V}, \mathbf{h}^{(1)}, \mathbf{H}^{(2)}; \boldsymbol{\theta})\right)}{\mathcal{Z}(\boldsymbol{\theta}, N)},$$

---
[1] This model can also be seen as a Dual-Wing Harmonium [17] in which one wing is unclamped.

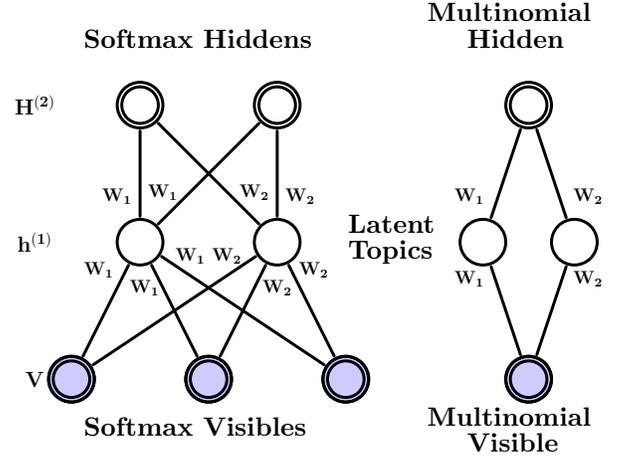

Figure 2: The Over-Replicated Softmax model. The bottom layer represents softmax visible units $\mathbf{V}$. The middle layer represents binary latent topics $\mathbf{h}^{(1)}$. The top layer represents softmax hidden units $\mathbf{H}^{(2)}$. All visible and hidden softmax units share the same set of weights, connecting them to binary hidden units. **Left:** The model for a document containing $N = 3$ words with $M = 2$ softmax hidden units. **Right:** A different interpretation of the model, in which $N$ softmax units with identical weights are replaced by a single multinomial unit which is sampled $N$ times and the $M$ softmax hidden units are replaced by a multinomial unit sampled $M$ times.

Note that the normalizing constant depends on the number of words N in the corresponding document, since the model contains as many visible softmax units as there are words in the document. So the model can be viewed as a family of different-sized DBMs that are created for documents of different lengths, but with a fixed-sized second-layer.

A pleasing property of the Over-Replicated Softmax model is that it has exactly the same number of trainable parameters as the Replicated Softmax model. However, the model's marginal distribution over $\mathbf{V}$ is different, as the second hidden layer provides an additional implicit prior. The model's prior over the latent topics $\mathbf{h}^{(1)}$ can be viewed as the geometric mean of the two probability distributions: one defined by an RBM composed of $\mathbf{V}$ and $\mathbf{h}^{(1)}$, and the other defined by an RBM composed of $\mathbf{h}^{(1)}$ and $\mathbf{H}^{(2)}$:[2]

$$P(\mathbf{h}^{(1)}; \boldsymbol{\theta}) = \frac{1}{\mathcal{Z}(\boldsymbol{\theta}, N)} \underbrace{\left(\sum_{\mathbf{v}} \exp\left(\sum_{j=1}^{F}\sum_{k=1}^{K} W_{jk} \hat{v}_k h_j^{(1)}\right)\right)}_{\text{RBM with } \mathbf{h}^{(1)} \text{ and } \mathbf{v}}$$
$$\underbrace{\left(\sum_{\mathbf{H}^{(2)}} \exp\left(\sum_{j=1}^{F}\sum_{k=1}^{K} W_{jk} \hat{h}_k^{(2)} h_j^{(1)}\right)\right)}_{\text{RBM with } \mathbf{h}^{(1)} \text{ and } \mathbf{H}^{(2)}}. \quad (7)$$

Observe that $\sum_{k=1}^{K} \hat{v}_k = N$ and $\sum_{k=1}^{K} \hat{h}_k^{(2)} = M$, so the strength of this prior can be varied by changing the number $M$ of second-layer softmax units. For example, if $M = N$, then the model's marginal distribution over $\mathbf{h}^{(1)}$, defined in Eq. 7, is given by the product of two identical distributions. In this DBM, the second-layer performs 1/2 of the modeling work compared to the first layer [11]. Hence, for documents containing few words ($N \ll M$) the prior over hidden topics $\mathbf{h}^{(1)}$ will be dominated by the second-layer, whereas for long documents ($N \gg M$) the effect of having a second-layer will diminish. As we show in our experimental results, having this additional flexibility in terms of defining an implicit prior over $\mathbf{h}^{(1)}$ significantly improves model performance, particularly for small and medium-sized documents.

---
[2] We omit the bias terms for clarity of presentation.

### 3.1 Learning

Let $\mathbf{h} = \{\mathbf{h}^{(1)}, \mathbf{H}^{(2)}\}$ be the set of hidden units in the two-layer DBM. Given a collection of L documents $\{\mathbf{V}\}_{l=1}^{L}$, the derivative of the log-likelihood with respect to model parameters $W$ takes the form:

$$\frac{1}{L}\sum_{l=1}^{L} \frac{\partial \log P(\mathbf{V_l}; \boldsymbol{\theta})}{\partial W_{jk}} = \mathbb{E}_{P_{\text{data}}}\left[(\hat{v}_k + \hat{h}_k^{(2)})h_j^{(1)}\right] - \mathbb{E}_{P_{\text{Model}}}\left[(\hat{v}_k + \hat{h}_k^{(2)})h_j^{(1)}\right],$$

where $\mathbb{E}_{P_{\text{data}}}[\cdot]$ denotes an expectation with respect to the data distribution $P_{\text{data}}(\mathbf{h}, \mathbf{V}) = P(\mathbf{h}|\mathbf{V}; \boldsymbol{\theta})P_{\text{data}}(\mathbf{V})$, with $P_{\text{data}}(\mathbf{V}) = \frac{1}{L}\sum_l \delta(\mathbf{V} - \mathbf{V}_l)$ representing the empirical distribution, and $\mathbb{E}_{P_{\text{Model}}}[\cdot]$ is an expectation with respect to the distribution defined by the model. Similar to the Replicated Softmax model, exact maximum likelihood learning is intractable, but approximate learning can be performed using a variational approach [10]. We use mean-field inference to estimate data-dependent expectations and an MCMC based stochastic approximation procedure to approximate the models expected sufficient statistics.

Consider any approximating distribution $Q(\mathbf{h}|\mathbf{V}; \boldsymbol{\mu})$, parameterized by a vector of parameters $\boldsymbol{\mu}$, for the posterior $P(\mathbf{h}|\mathbf{V}; \boldsymbol{\theta})$. Then the log-likelihood of the DBM model has the following variational lower bound:

$$\log P(\mathbf{V}; \boldsymbol{\theta}) \geq \sum_{\mathbf{h}} Q(\mathbf{h}|\mathbf{V}; \boldsymbol{\mu}) \log P(\mathbf{V}, \mathbf{h}; \boldsymbol{\theta}) + \mathcal{H}(Q),$$

where $\mathcal{H}(\cdot)$ is the entropy functional. The bound becomes tight if and only if $Q(\mathbf{h}|\mathbf{V}; \boldsymbol{\mu}) = P(\mathbf{h}|\mathbf{V}; \boldsymbol{\theta})$.

For simplicity and speed, we approximate the true posterior $P(\mathbf{h}|\mathbf{V}; \boldsymbol{\theta})$ with a fully factorized approximating distribution over the two sets of hidden units, which corresponds to the so-called mean-field approximation:

$$Q^{MF}(\mathbf{h}|\mathbf{V}; \boldsymbol{\mu}) = \prod_{j=1}^{F} q(h_j^{(1)}|\mathbf{V}) \prod_{i=1}^{M} q(h_i^{(2)}|\mathbf{V}), \quad (8)$$

where $\boldsymbol{\mu} = \{\boldsymbol{\mu}^{(1)}, \boldsymbol{\mu}^{(2)}\}$ are the mean-field parameters with $q(h_j^{(1)} = 1) = \mu_j^{(1)}$ and $q(h_{ik}^{(2)} = 1) = \mu_k^{(2)}$, $\forall i \in \{1, \ldots, M\}$, s.t. $\sum_{k=1}^{K} \mu_k^{(2)} = 1$. Note that due to the shared weights across all of the hidden softmaxes, $q(h_{ik}^{(2)})$ does not dependent on $i$. In this case, the variational lower bound takes a particularly simple form:

$$\log P(\mathbf{V}; \boldsymbol{\theta}) \geq \sum_{\mathbf{h}} Q^{MF}(\mathbf{h}|\mathbf{V}; \boldsymbol{\mu}) \log P(\mathbf{V}, \mathbf{h}; \boldsymbol{\theta}) + \mathcal{H}(Q^{MF})$$
$$\geq \left(\hat{\mathbf{v}}^\top + M\boldsymbol{\mu}^{(2)\top}\right) \mathbf{W}\boldsymbol{\mu}^{(1)} - \log \mathcal{Z}(\boldsymbol{\theta}, N) + \mathcal{H}(Q^{MF}),$$

where $\hat{\mathbf{v}}$ is a $K \times 1$ vector, with its $k^{th}$ element $\hat{v}_k$ containing the count for the $k^{th}$ word. Since $\sum_{k=1}^{K} \hat{v}_k = N$ and $\sum_{k=1}^{K} \mu_k^{(2)} = 1$, the first term in the bound linearly combines the effect of the data (which scales as $N$) with the prior (which scales as $M$). For each training example, we maximize this lower bound with respect to the variational parameters $\boldsymbol{\mu}$ for fixed parameters $\boldsymbol{\theta}$, which results in the mean-field fixed-point equations:

$$\mu_j^{(1)} \leftarrow \sigma\left(\sum_{k=1}^{K} W_{jk}\left(\hat{v}_k + M\mu_k^{(2)}\right)\right), \quad (9)$$

$$\mu_k^{(2)} \leftarrow \frac{\exp\left(\sum_{j=1}^{F} W_{jk}\mu_j^{(1)}\right)}{\sum_{k'=1}^{K} \exp\left(\sum_{j=1}^{F} W_{jk'}\mu_j^{(1)}\right)}, \quad (10)$$

where $\sigma(x) = 1/(1 + \exp(-x))$ is the logistic function. To solve these fixed-point equations, we simply cycle through layers, updating the mean-field parameters within a single layer.

Given the variational parameters $\boldsymbol{\mu}$, the model parameters $\boldsymbol{\theta}$ are then updated to maximize the variational bound using an MCMC-based stochastic approximation [10, 15, 18]. Let $\boldsymbol{\theta}_t$ and $\mathbf{x}_t = \{\mathbf{V}_t, \mathbf{h}^{(1)}_t, \mathbf{h}^{(2)}_t\}$ be the current parameters and the state. Then $\mathbf{x}_t$ and $\boldsymbol{\theta}_t$ are updated sequentially as follows: given $\mathbf{x}_t$, sample a new state $\mathbf{x}_{t+1}$ using alternating Gibbs sampling. A new parameter $\boldsymbol{\theta}_{t+1}$ is then obtained by making a gradient step, where the intractable model's expectation $\mathbb{E}_{P_{\text{model}}}[\cdot]$ in the gradient is replaced by a point estimate at sample $\mathbf{x}_{t+1}$.

In practice, to deal with variable document lengths, we take a minibatch of data and run one Markov chain for each training case for a few steps. To update the model parameters, we use an average over those chains. Similar to Contrastive Divergence learning, in order to provide a good starting point for the sampling, we initialize each chain at $\hat{\mathbf{h}}^{(1)}$ by sampling from the mean-field approximation to the posterior $q(\mathbf{h}^{(1)}|\mathbf{V})$.

### 3.2 An Efficient Pretraining Algorithm

The proper training procedure for the DBM model described above is quite slow. This makes it very impor-tant to pretrain the model so that the model parameters start off in a nice region of space. Fortunately, due to parameter sharing between the visible and hidden softmax units, there exists an efficient pretraining method which makes the proper training almost redundant.

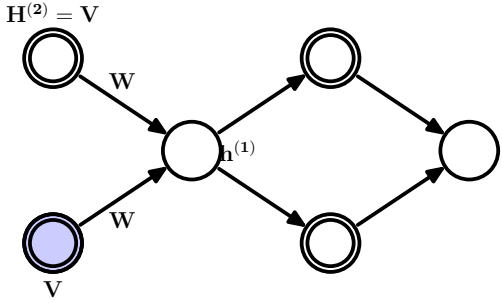

Figure 3: Pretraining a two-layer Boltzmann Machine using one-step contrastive divergence. The second hidden softmax layer is initialized to be the same as the observed data. The units in the first hidden layer have stochastic binary states, but the reconstructions of both the visible and second hidden layer use probabilities, so both reconstructions are identical.

Consider a DBM with $N$ observed and $M$ hidden softmax units. Let us first assume that the number of hidden softmaxes $M$ is the same as the number of words $N$ in a given document. If we were given the initial state vector $\mathbf{H}^{(2)}$, we could train this DBM using one-step contrastive divergence with mean-field reconstructions of both the states of the visible and the hidden softmax units, as shown in Fig. 3. Since we are not given the initial state, one option is to set $\mathbf{H}^{(2)}$ to be equal to the data $\mathbf{V}$. Provided we use mean-field reconstructions for both the visible and second-layer hidden units, one-step contrastive divergence is then exactly the same as training a Replicated Softmax RBM with only one hidden layer but with bottom-up weights that are twice the top-down weights.

To pretrain a DBM with different number of visible and hidden softmaxes, we train an RBM with the bottom-up weights scaled by a factor of $1 + \frac{M}{N}$. In other words, in place of using $\mathbf{W}$ to compute the conditional probability of the hidden units (see Eq. 3), we use $(1 + \frac{M}{N})\mathbf{W}$:

$$P(h_j^{(1)} = 1|\mathbf{V}) = \sigma\left((1 + \frac{M}{N})\sum_{k=1}^{K} v_k W_{kj}\right). \quad (11)$$

The conditional probability of the observed softmax units remains the same as in Eq. 4. This procedure is equivalent to training an RBM with $N + M$ observed visible units with each of the $M$ extra units set to be the empirical word distribution in the document, i.e..

for $i \in \{N+1, \ldots, N+M\}$,

$$v_{ik} = \frac{\sum_{j=1}^{N} v_{jk}}{\sum_{j=1}^{N} \sum_{k'=1}^{K} v_{jk'}}$$

Thus the $M$ extra units are not 1-of-K, but represent distributions over the $K$ words[3].

This way of pretraining the Over-Replicated Softmax DBMs with tied weights will not in general maximize the likelihood of the weights. However, in practice it produces models that reconstruct the training data well and serve as a good starting point for generative fine-tuning of the two-layer model.

### 3.3 Inference

The posterior distribution $P(\mathbf{h}^{(1)}|\mathbf{V})$ represents the latent topic structure of the observed document. Conditioned on the document, these activation probabilities can be inferred using the mean-field approximation used to infer data-dependent statistics during training.

A fast alternative to the mean-field posterior is to multiply the visible to hidden weights by a factor of $1 + \frac{M}{N}$ and approximate the true posterior with a single matrix multiply, using Eq. 11. Setting $M = 0$ recovers the proper posterior inference step for the standard Replicated Softmax model. This simple scaling operation leads to significant improvements. The results reported for retrieval and classification experiments used the fast pretraining and fast inference methods.

### 3.4 Choosing $M$

The number of hidden softmaxes $M$ affects the strength of the additional prior. The value of $M$ can be chosen using a validation set. Since the value of $M$ is fixed for all Over-Replicated DBMs, the effect of the prior will be less for documents containing many words. This is particularly easy to see in Eq. 11. As $N$ becomes large, the scaling factor approaches 1, diminishing the part of implicit prior coming from the $M$ hidden softmax units. Thus the value of $M$ can be chosen based on the distribution of lengths of documents in the corpus.

## 4 Experiments

In this section, we evaluate the Over-Replicated Softmax model both as a generative model and as a feature extraction method for retrieval and classification. Two datasets are used - 20 Newsgroups and Reuters Corpus Volume I (RCV1-v2).

---
[3]Note that when $M = N$, we recover the setting of having the bottom-up weights being twice the top-down weights.

### 4.1 Description of datasets

The 20 Newsgroups dataset consists of 18,845 posts taken from the Usenet newsgroup collection. Each post belongs to exactly one newsgroup. Following the preprocessing in [12] and [7], the data was partitioned chronologically into 11,314 training and 7,531 test articles. After removing stopwords and stemming, the 2000 most frequent words in the training set were used to represent the documents.

The Reuters RCV1-v2 contains 804,414 newswire articles. There are 103 topics which form a tree hierarchy. Thus documents typically have multiple labels. The data was randomly split into 794,414 training and 10,000 test cases. The available data was already preprocessed by removing common stopwords and stemming. We use a vocabulary of the 10,000 most frequent words in the training dataset.

### 4.2 Training details

The Over-Replicated Softmax model was first pretrained with Contrastive Divergence using the weight scaling technique described in Sec. 3.2. Minibatches of size 128 were used. A validation set was held out from the training set for hyperparameter selection (1,000 cases for 20 newsgroups and 10,000 for RCV1-v2). The value of $M$ and number of hidden units were chosen over a coarse grid using the validation set. Typically, $M = 100$ performed well on both datasets. Increasing the number of hidden units lead to better performance on retrieval and classification tasks, until serious overfitting became a problem around 1000 hidden units. For perplexity, 128 hidden units worked quite well and having too many units made the estimates of the partition function obtained using AIS unstable. Starting with CD-1, the number of Gibbs steps was stepped up by one after every 10,000 weight updates till CD-20. Weight decay was used to prevent overfitting. Additionally, in order to encourage sparsity in the hidden units, KL-sparsity regularization was used. We decayed the learning rate as $\frac{\epsilon_0}{1+t/T}$, with $T = 10,000$ updates. This approximate training was sufficient to give good results on retrieval and classification tasks. However, to obtain good perplexity results, the model was trained properly using the method described in Sec. 3.1. Using 5 steps for mean-field inference and 20 for Gibbs sampling was found to be sufficient. This additional training gave improvements in terms of perplexity but the improvement on classification and retrieval tasks was not statistically significant.

We also implemented the standard Replicated Softmax model. The training procedure was the same as the pretraining process for the Over-Replicated Softmax model. Both the models were implemented on GPUs. Pretraining took 3-4 hours for the 2-layered Boltzmann Machines (depending on $M$) and the proper training took 10-12 hours. The DocNADE model was run using the publicly available code[4]. We used default settings for all hyperparameters, except the learning rates which were tuned separately for each hidden layer size and data set.

Table 1: Comparison of the average test perplexity per word. All models use 128 topics.

|  | 20 News | Reuters |
|---|---|---|
| Training set size | 11,072 | 794,414 |
| Test set size | 7,052 | 10,000 |
| Vocabulary size | 2,000 | 10,000 |
| Avg Document Length | 51.8 | 94.6 |
| **Perplexities** | | |
| Unigram | 1335 | 2208 |
| Replicated Softmax | 965 | 1081 |
| Over-Rep. Softmax ($M = 50$) | 961 | 1076 |
| Over-Rep. Softmax ($M = 100$) | **958** | **1060** |

### 4.3 Perplexity

We compare the Over-Replicated Softmax model with the Replicated Softmax model in terms of perplexity. Computing perplexities involves computing the partition functions for these models. We used Annealed Importance Sampling [9] for doing this. In order to get reliable estimates, we ran 128 Markov chains for each document length. The average test perplexity per word was computed as $\exp\left(-1/L \sum_{l=1}^{L} 1/N_l \log p(\mathbf{v}_l)\right)$, where $N_l$ is the number of words in document $l$. Table 1 shows the perplexity averaged over $L = 1000$ randomly chosen test cases for each data set. Each of the models has 128 latent topics. Table 1 shows that the Over-Replicated Softmax model assigns slightly lower perplexity to the test data compared to the Replicated Softmax model. For the Reuters data set the perplexity decreases from 1081 to 1060, and for 20 Newsgroups, it decreases from 965 to 958. Though the decrease is small, it is statistically significant since the standard deviation was typically ±2 over 10 random choices of 1000 test cases. Increasing the value of $M$ increases the strength of the prior, which leads to further improvements in perplexities. Note that the estimate of the log probability for 2-layered Boltzmann Machines is a lower bound on the actual log probability. So the perplexities we show are upper bounds and the actual perplexities may be lower (provided the estimate of the partition function is close to the actual value).

---

[4] http://www.dmi.usherb.ca/~larocheh/code/DocNADE.zip

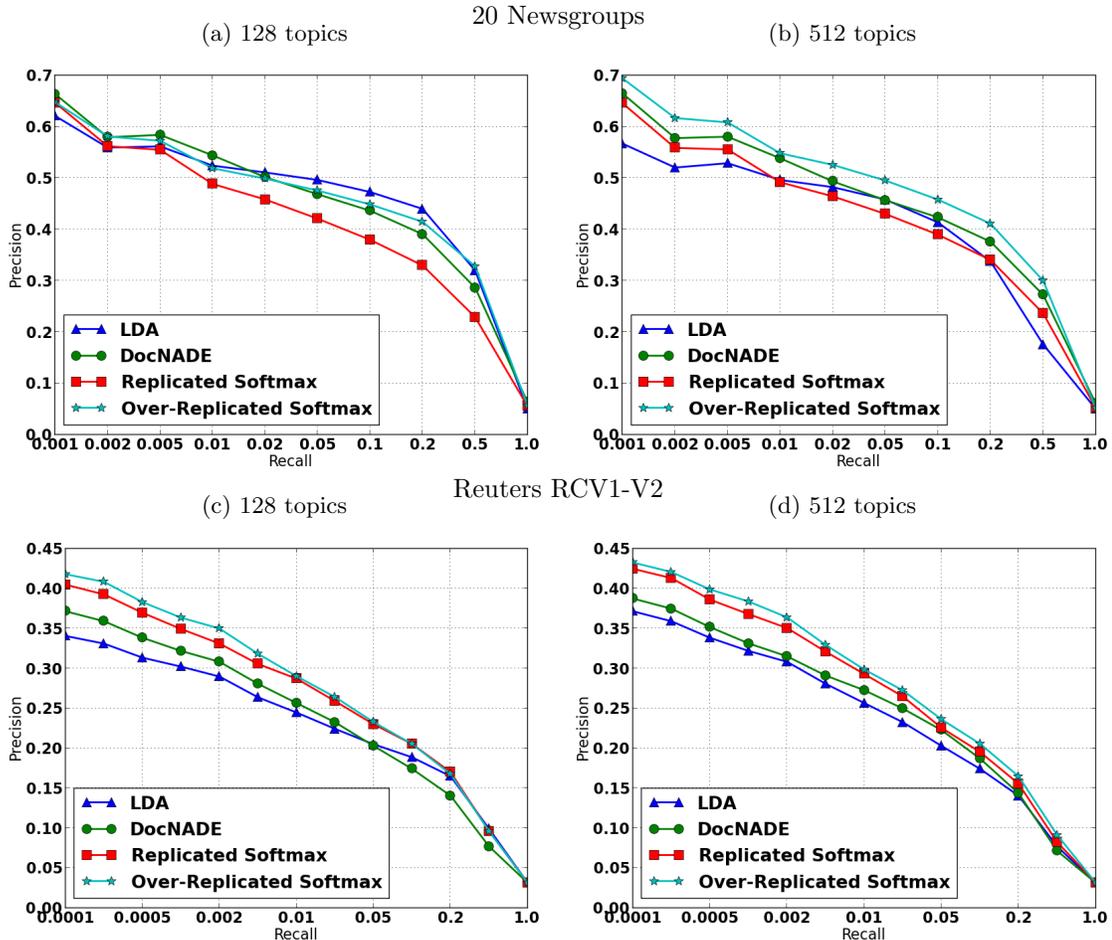

Figure 4: Comparison of Precision-Recall curves for document retrieval. All Over-Replicated Softmax models use $M = 100$ latent words.

### 4.4 Document Retrieval

In order to do retrieval, we represent each document $\mathbf{V}$ as the conditional posterior distribution $P(\mathbf{h}^{(1)}|\mathbf{V})$. This can be done exactly for the Replicated Softmax and DocNADE models. For two-layered Boltzmann Machines, we extract this representation using the fast approximate inference as described in Sec. 3.3. Performing more accurate inference using the mean-field approximation method did not lead to statistically different results. For the LDA, we used 1000 Gibbs sweeps per test document in order to get an approximate posterior over the topics.

Documents in the training set (including the validation set) were used as a database. The test set was used as queries. For each query, documents in the database were ranked using cosine distance as the similarity metric. The retrieval task was performed separately for each label and the results were averaged. Fig. 4 compares the precision-recall curves. As shown by Fig. 4, the Over-Replicated Softmax DBM out- performs other models on both datasets, particularly when retrieving the top few documents.

To find the source of improvement, we analyzed the effect of document length of retrieval performance. Fig. 5 plots the average precision obtained for query documents arranged in order of increasing length. We found that the Over-Replicated Softmax model gives large gains on documents with small numbers of words, confirming that the implicit prior imposed using a fixed value of $M$ has a stronger effect on short documents. As shown in Fig. 5, DocNADE and Replicated Softmax models often do not do well for documents with few words. On the other hand, the Over-Replicated softmax model performs significantly better for short documents. In most document collections, the length of documents obeys a power law distribution. For example, in the 20 newsgroups dataset 50% of the documents have fewer than 35 words (Fig. 5c). This makes it very important to do well on short documents. The Over-Replicated Softmax model achieves this goal.

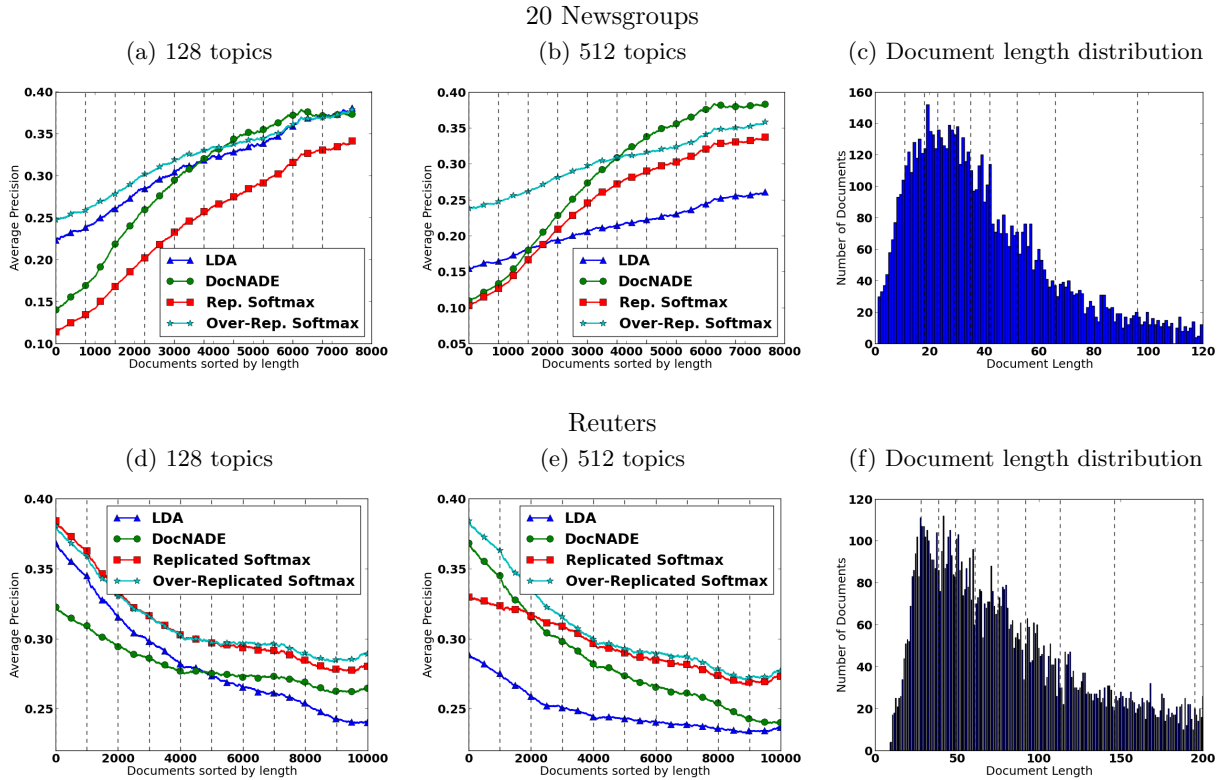

Figure 5: Effect of document size on retrieval performance for different topic models. The x-axis in Figures (a), (b), (d), (e) represents test documents arranged in increasing order of their length. The y-axis shows the average precision obtained by querying that document. The plots were smoothed to make the general trend visible. Figures (c) and (f) show the histogram of document lengths for the respective datasets. The dashed vertical lines denote 10-percentile boundaries. **Top :** Average Precision on the 20 Newsgroups dataset. **Bottom :** Mean Average Precision on the Reuters dataset. The Over-Replicated Softmax models performs significantly better for documents with few words. The adjoining histograms in each row show that such documents occur quite frequently in both data sets.

### 4.5 Document Classification

In this set of experiments, we evaluate the learned representations from the Over-Replicated Softmax model for the purpose of document classification. Since the objective is to evaluate the quality of the representation, simple linear classifiers were used. Multinomial logistic regression with a cross entropy loss function was used for the 20 newsgroups data set. The evaluation metric was classification accuracy. For the Reuters dataset, we used independent logistic regressions for each label since it is a multi-label classification problem. The evaluation metric was Mean Average Precision.

Table 2 shows the results of these experiments. The Over-Replicated Softmax model performs significantly better than the standard Replicated Softmax model and LDA across different network sizes on both datasets. For the 20 newsgroups dataset using 512 topics, LDA gets 64.2% accuracy. Replicated Softmax (67.7%) and DocNADE (68.4%) improve upon this. The Over-Replicated Softmax model further improves the result to 69.4%. The difference is larger for the Reuters dataset. In terms of Mean Average Precision (MAP), the Over-Replicated Softmax model achieves 0.453 which is a very significant improvement upon DocNADE (0.427) and Replicated Softmax (0.421).

Table 2: Comparison of Classification accuracy on 20 Newsgroups dataset and Mean Average Precision on Reuters RCV1-v2.

| Model | 20 News | | Reuters | |
| --- | --- | --- | --- | --- |
| | 128 | 512 | 128 | 512 |
| LDA | 65.7 | 64.2 | 0.304 | 0.351 |
| DocNADE | **67.0** | 68.4 | 0.388 | 0.417 |
| Replicated Softmax | 65.9 | 67.7 | 0.390 | 0.421 |
| Over-Rep. Softmax | 66.8 | **69.1** | **0.401** | **0.453** |

We further examined the source of improvement by analyzing the effect of document length on the classification performance. Similar to retrieval, we found that the Over-Replicated Softmax model performs well on short documents. For long documents, the performance of the different models was similar.

## 5 Conclusion

The Over-Replicated Softmax model described in this paper is an effective way of defining a flexible prior over the latent topic features of an RBM. This model causes no increase in the number of trainable parameters and only a minor increase in training algorithm complexity. Deep Boltzmann Machines are typically slow to train. However, our fast approximate training method makes it possible to train the model with CD, just like an RBM. The features extracted from documents using the Over-Replicated Softmax model perform better than features from the standard Replicated Softmax and LDA models and are comparable to DocNADE across different network sizes.

While the number of hidden softmax units $M$, controlling the strength of the prior, was chosen once and fixed across all DBMs, it is possible to have $M$ depend on $N$. One option is to set $M = cN$, $c > 0$. In this case, for documents of all lengths, the second-layer would perform perform $c/c+1$ of the modeling work compared to the first layer. Another alternative is to set $M = N_{max} - N$, where $N_{max}$ is the maximum allowed length of all documents. In this case, our DBM model will always have the same number of replicated softmax units $N_{max} = N + M$, hence the same architecture and a single partition function. Given a document of length N, the remaining $N_{max} - N$ words can be treated as missing. All of these variations improve upon the standard Replicated Softmax model, LDA, and DocNADE models, opening up the space of new deep undirected topics to explore.